\documentclass[pdflatex,sn-nature]{sn-jnl}

\usepackage{graphicx}%
\usepackage{multirow}%
\usepackage{amsmath,amssymb,amsfonts}%
\usepackage{amsthm}%
\usepackage{mathrsfs}%
\usepackage[title]{appendix}%
\usepackage{xcolor}%
\usepackage{textcomp}%
\usepackage{manyfoot}%
\usepackage{booktabs}%
\usepackage{algorithm}%
\usepackage{algorithmicx}%
\usepackage{algpseudocode}%
\usepackage{listings}%

\theoremstyle{thmstyleone}%
\theoremstyle{thmstyletwo}%

\theoremstyle{thmstylethree}%

\begin{document}

\title[ChlCast]{An AI-enabled high-resolution chlorophyll-a forecasting model that outperforms the operational global biogeochemical system}


\author[1]{\fnm{Yong} \sur{Jiang}}

\author[1]{\fnm{Zheng} \sur{Jiang}}

\author[1]{\fnm{Gaowei} \sur{Zhang}}

\author[3]{\fnm{Suixiang} \sur{Shi}}

\author[4]{\fnm{Zengzhou} \sur{Hao}}

\author*[2]{\fnm{Chi} \sur{Zhang}}\email{zhangchi2@pjlab.org.cn}

\author*[1]{\fnm{Wei} \sur{Wang}}\email{weiwang@bupt.edu.cn}

\affil*[1]{
\orgdiv{School of Artificial Intelligence},
\orgname{Beijing University of Posts and Telecommunications},
\orgaddress{
\city{Beijing},
\country{China}
}
}

\affil*[2]{
\orgname{Shanghai Artificial Intelligence Laboratory},
\orgaddress{
\city{Shanghai},
\country{China}
}
}

\affil[3]{
\orgdiv{Key Laboratory of Digital Ocean},
\orgname{National Marine Data and Information Service},
\orgaddress{
\city{Tianjin},
\country{China}
}
}

\affil[4]{
\orgdiv{State Key Laboratory of Satellite Ocean Environment Dynamics},
\orgname{Second Institute of Oceanography, Ministry of Natural Resources},
\orgaddress{
\city{Hangzhou},
\country{China}
}
}


\abstract{Surface chlorophyll-a concentration is a key indicator of marine phytoplankton variability, and short-term forecasts of its distribution provide advance information for marine ecosystem monitoring and management. However, interactions between physical transport and biological processes complicate chlorophyll-a forecasting, and simulating these coupled processes with conventional numerical models imposes substantial computational demands. Here we introduce ChlCast, an AI model that generates daily forecasts of surface chlorophyll-a concentration over a 10-day horizon at approximately 4-km resolution between 60$^\circ$S and 60$^\circ$N. ChlCast combines factorized spatiotemporal convolutions with a multiscale architecture to learn from historical chlorophyll-a fields and climatological information. Compared with the operational 0.25$^\circ$ Copernicus global ocean biogeochemical forecasting system, ChlCast reduces mean absolute error and root mean square error by 70.74\% and 47.63\%, respectively, against satellite-derived Level-4 chlorophyll-a fields, and by 35.03\% and 13.07\% against independent BGC-Argo in situ observations. Coastal case studies further show that the 4-km forecasts retain coastal gradients and localized concentration maxima that are weakened by coarse-grid averaging. These results demonstrate the potential of observation-driven deep learning for high-resolution forecasting of surface chlorophyll-a variability at the near-global scale.}




\maketitle
\section{Introduction}\label{sec1}

Phytoplankton are the primary producers in marine ecosystems, sustaining marine food webs and contributing to the global carbon cycle through the biological carbon pump~\cite{field1998,falkowski1998}.
These contributions vary with phytoplankton abundance and distribution, which are heterogeneous across the global ocean and change over time~\cite{yoder2003,behrenfeld2006}. 
Understanding this variability is therefore important for assessing the ecological and biogeochemical roles of phytoplankton across regions and timescales~\cite{siegel2014}. 
Surface chlorophyll-a concentration is widely used as an indicator of phytoplankton biomass, providing an observational basis for characterizing phytoplankton variability in the surface ocean~\cite{lim2022,mcclain2009}. 
Short-term forecasts complement observations by anticipating how surface chlorophyll-a distributions may evolve over the coming days~\cite{lazzari2010,teruzzi2014}. 
Such advance information can guide monitoring efforts and support marine ecosystem management~\cite{gehlen2015,brasseur2009}. 
Supporting these applications requires forecasts that capture fine-scale chlorophyll-a variability, which may be obscured on coarser grids~\cite{levy2018,fennel2022}.

However, accurate short-term forecasting of surface chlorophyll-a remains challenging because its evolution depends on interactions between physical transport and biological processes. 
Ocean circulation and mixing shape surface chlorophyll-a variability by redistributing phytoplankton biomass and altering light and nutrient availability~\cite{aumont2015,chavez2011}. 
The nonlinear nature of these interactions and their varying importance across regions and seasons complicate prediction at the global scale~\cite{feng2015,behrenfeld2014}. 
To enable short-term chlorophyll-a forecasting, conventional numerical systems, such as the Copernicus global ocean biogeochemical forecasting system, couple ocean circulation models with biogeochemical models. These systems predict chlorophyll-a concentrations by numerically integrating the equations governing these coupled processes~\cite{copernicus_quality,butenschon2016}. 
Despite their predictive capabilities, forecast accuracy remains sensitive to uncertainties in model parameterizations, such as those describing biological rates and responses to light and nutrient availability~\cite{schartau2017,friedrichs2007}. 
In addition, running these coupled simulations over extensive ocean regions entails substantial computational costs, often limiting the spatial resolution of operational global forecasts~\cite{holt2017,hewitt2017}.

Recent advances in deep learning have enabled effective modelling of complex nonlinear spatiotemporal relationships~\cite{lecun2015,reichstein2019}. 
Following progress in global weather forecasting~\cite{lam2023}, data-driven models have increasingly been developed to predict physical ocean variables, including temperature, salinity and currents, demonstrating the potential for accurate and computationally efficient daily forecasts~\cite{cui2025}. 
Existing studies have begun to explore data-driven chlorophyll-a forecasting in regional settings (e.g., the Pearl River Estuary and the northern Gulf of Mexico) or at monthly to seasonal timescales~\cite{zhang2025stimp,martinez2026}. 
However, daily, high-resolution forecasting across the near-global ocean remains insufficiently explored. 
Achieving such forecasts presents two main challenges. 
First, daily forecasting requires capturing multiscale spatial structures and their day-to-day evolution while accounting for regionally varying seasonal backgrounds~\cite{mahadevan2016,hong2023}. Second, preserving fine-scale chlorophyll-a variability across a near-global domain substantially increases both data volume and computational demands, requiring an efficient architecture that balances spatial detail and computational cost~\cite{bi2023,bodnar2025}.

Here, we introduce ChlCast, a deep learning model that generates daily forecasts of surface chlorophyll-a concentration over a 10-day horizon at approximately 4-km spatial resolution across a near-global domain spanning $60^\circ$S--$60^\circ$N. 
This domain covers most of the global ocean area and encompasses diverse chlorophyll-a regimes, ranging from oligotrophic subtropical gyres to productive coastal and upwelling waters~\cite{siegel2005}.
ChlCast combines factorized spatiotemporal convolutions with a multiscale encoder--decoder architecture to model spatial structure and temporal evolution from historical chlorophyll-a sequences and climatological background information. 
During the evaluation period, ChlCast achieved lower mean absolute error (MAE) and root mean square error (RMSE) and higher Pearson correlation coefficients than the operational Copernicus numerical forecasts when evaluated against both satellite-derived Level-4 chlorophyll-a fields and independent BGC-Argo observations, and consistently outperformed simple baselines across all lead times.
Case studies in the eastern Mediterranean, the Seto Inland Sea and the southwestern Baltic Sea further showed that the 4-km forecasts preserved nearshore gradients, localized concentration maxima and alongshore heterogeneity that were substantially weakened after aggregation to the $0.25^\circ$ grid.
A full-domain 10-day forecast required approximately 34~s on a single NVIDIA GeForce RTX 3090 GPU, demonstrating the potential of observation-driven deep learning for computationally efficient, high-resolution chlorophyll-a forecasting at the near-global scale.

\section{Results}\label{sec2}

\subsection{Comparison with the Copernicus operational forecasting system}\label{subsec1}

\begin{figure}[t]
    \centering
    \includegraphics[width=\linewidth]{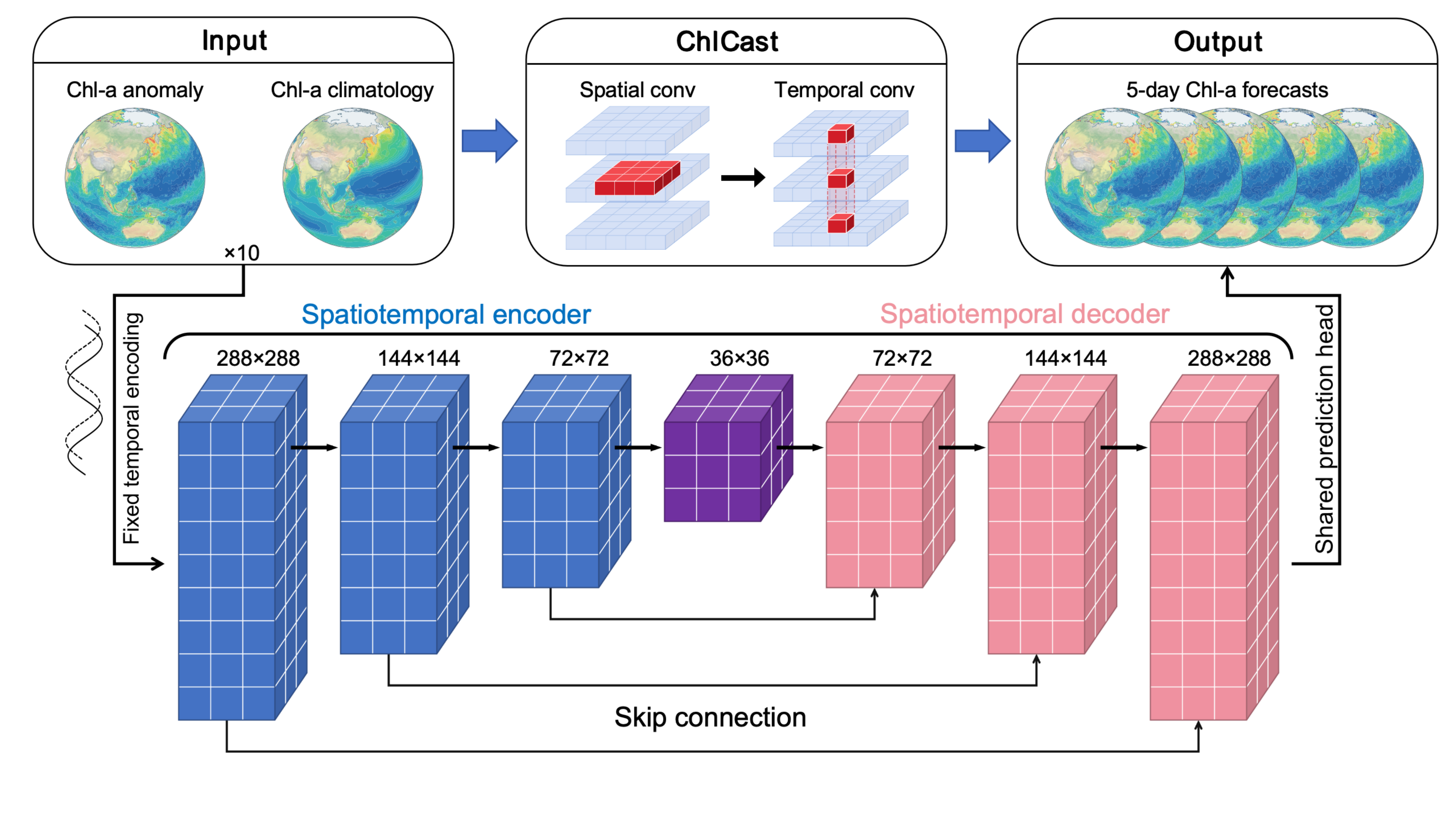}
    \caption{\textbf{Schematic of the ChlCast architecture.}
    ChlCast uses log-space chlorophyll-a anomalies and the corresponding
    log-transformed daily climatology over the preceding 10 days as a
    two-channel input. Fixed temporal encoding provides information on
    the position of each input day. The model adopts a (2+1)D U-Net
    architecture, in which separate spatial and temporal convolutions
    capture spatiotemporal dependencies. The encoder (blue) extracts
    features at progressively coarser spatial scales, whereas the
    decoder (pink) restores spatial resolution and incorporates features
    from the corresponding encoder levels through skip connections.
    Numbers indicate the spatial dimensions of the feature maps.
    ChlCast generates log-space anomaly forecasts for
    the subsequent 10 days, which are combined with the corresponding
    target-day climatology and inverse-transformed to recover daily
    chlorophyll-a concentrations.}
    \label{fig:chlcast_architecture}
\end{figure}

ChlCast is a U-Net deep learning model for forecasting surface chlorophyll-a concentration across the ocean between 60$^\circ\mathrm{S}$ and 60$^\circ\mathrm{N}$. Using a 10-day sequence of log-space chlorophyll-a anomalies and corresponding daily climatological fields, the model generates daily forecasts for the next 10 days at a spatial resolution of approximately 4 km.
The model uses factorized spatial and temporal convolutions within a multiscale encoder–decoder framework to capture the spatiotemporal evolution of chlorophyll-a fields. 
The architecture of ChlCast (Fig.~\ref{fig:chlcast_architecture}) and its training details are described in Methods.

The Copernicus Global Ocean Biogeochemistry Analysis and Forecast product served as the numerical benchmark for evaluating ChlCast. This operational product is updated daily and provides 10-day global forecasts of surface chlorophyll-a concentration at a horizontal resolution of $0.25^\circ\times0.25^\circ$~\cite{copernicus_bgc}.
ChlCast and Copernicus forecasts were evaluated at lead times of 1--10 days against satellite-derived L4 surface chlorophyll-a fields. Both the ChlCast forecasts and L4 reference fields were aggregated to the \(0.25^\circ \times 0.25^\circ\) Copernicus grid over \(60^\circ\mathrm{S}\)--\(60^\circ\mathrm{N}\) across 56 valid forecast initializations between 14 August and 9 October 2026. 

\begin{figure}[t]
    \centering
    \includegraphics[width=\textwidth]{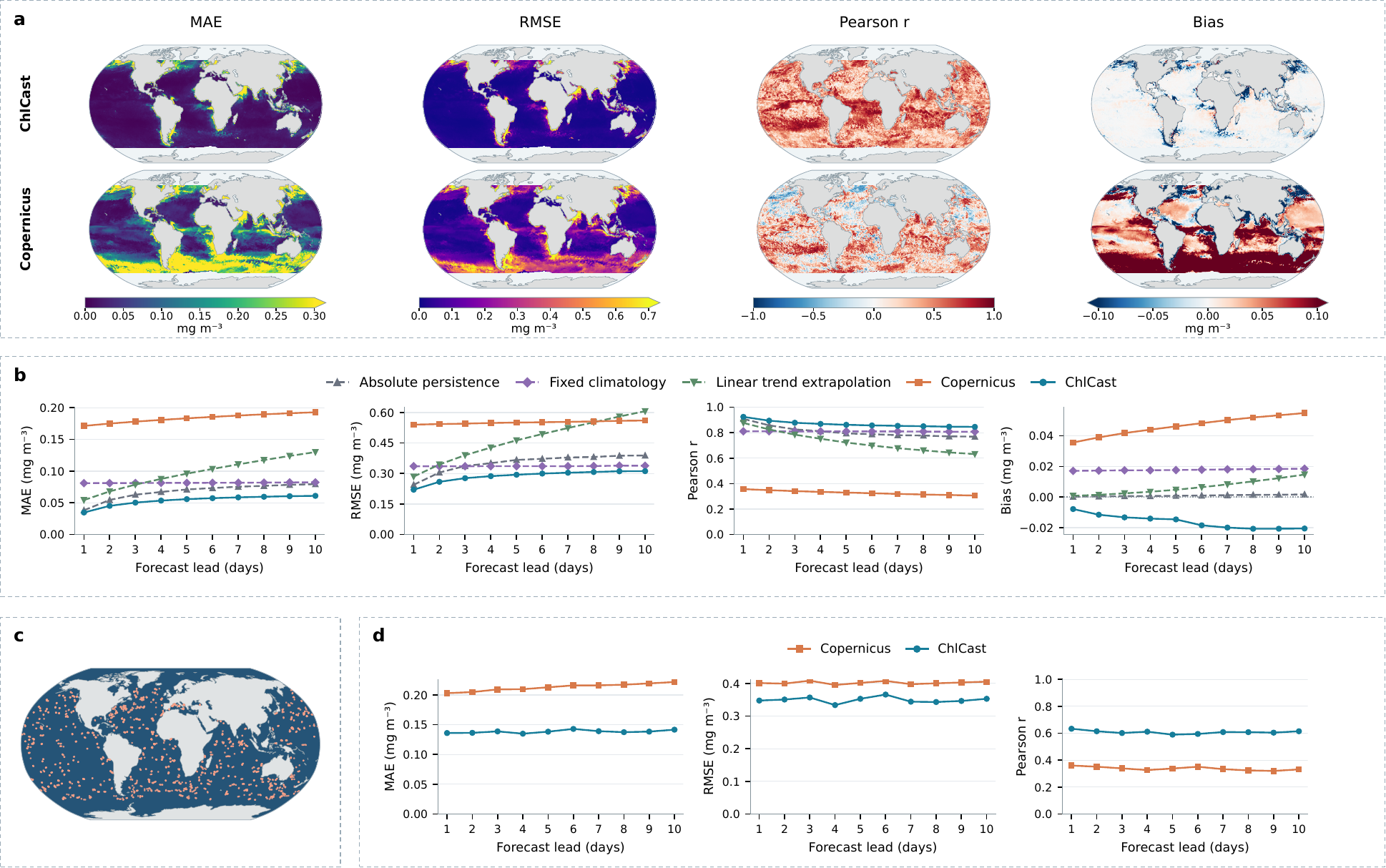}
\caption{\textbf{Comparison of ChlCast with the Copernicus operational global ocean biogeochemical forecasting system.}
    \textbf{a}, Spatial distributions of MAE, RMSE, Pearson correlation coefficient ($r$) and mean bias for ChlCast and Copernicus, evaluated against satellite-derived L4 surface chlorophyll-a fields over $60^\circ\mathrm{S}$--$60^\circ\mathrm{N}$.
    \textbf{b}, Forecast performance against the satellite L4 reference fields as a function of lead time for ChlCast, Copernicus and three baselines (absolute persistence, fixed climatology and linear trend extrapolation). Metrics include MAE, RMSE, Pearson $r$ and mean bias.
    \textbf{c}, Spatial distribution of the BGC-Argo profiles used for independent in situ validation. The validation dataset comprises 1,915 profiles from 505 floats.
    \textbf{d}, MAE, RMSE and Pearson $r$ of ChlCast and Copernicus at forecast lead times of 1--10 days, evaluated against BGC-Argo surface chlorophyll-a observations.
    MAE, RMSE and mean bias are expressed in $\mathrm{mg\,m^{-3}}$; $r$ is dimensionless.
    }
    \label{fig:chlcast_copernicus}
\end{figure}

Overall, ChlCast outperformed Copernicus across most of the evaluated domain (Fig.~\ref{fig:chlcast_copernicus}a). The advantage of ChlCast was particularly pronounced in the mid-latitude Southern Hemisphere, where Copernicus exhibited zonally extensive bands of elevated MAE and RMSE that were substantially reduced in ChlCast. ChlCast also showed higher Pearson correlations over much of the ocean, including areas where Copernicus correlations were low or negative. Moreover, Copernicus exhibited broad regions of positive and negative bias, whereas ChlCast biases remained closer to zero over much of the open ocean.

When data from all ten forecast days were pooled, ChlCast achieved an MAE of $0.0537~\mathrm{mg\,m^{-3}}$ and an RMSE of $0.2887~\mathrm{mg\,m^{-3}}$, corresponding to reductions of 70.74\% and 47.63\%, respectively, relative to Copernicus ($0.1836$ and $0.5512~\mathrm{mg\,m^{-3}}$). The pooled Pearson correlation coefficient was 0.8685 for ChlCast, compared with 0.3283 for Copernicus. Evaluation at individual forecast lead times further showed that ChlCast achieved lower MAE and RMSE and higher Pearson correlation than Copernicus across all lead days 1--10 (Fig.~\ref{fig:chlcast_copernicus}b). As expected, ChlCast errors increased with lead time: from day 1 to day 10, MAE rose from 0.0348 to $0.0611~\mathrm{mg\,m^{-3}}$, RMSE increased from 0.2205 to $0.3119~\mathrm{mg\,m^{-3}}$, and the Pearson correlation coefficient decreased from 0.9247 to 0.8460. Correspondingly, the relative advantage of ChlCast over Copernicus narrowed with lead time, with the reductions in MAE and RMSE decreasing from 79.70\% to 68.33\% and from 59.20\% to 44.36\%, respectively, between day 1 and day 10.
ChlCast also consistently outperformed forecasts based on unchanged initial conditions (absolute persistence), a fixed climatological background (fixed climatology) and extrapolation of recent trends (linear trend extrapolation) in MAE, RMSE and Pearson correlation across all evaluated lead times.

The bias results showed that ChlCast consistently underestimated chlorophyll-a concentrations at all forecast lead times, whereas Copernicus exhibited an overall positive bias. Absolute persistence and linear trend extrapolation had mean biases closer to zero. However, because global mean bias can mask compensating regional over- and underestimations, it should not be interpreted in isolation. The error characteristics are therefore better assessed together with MAE and RMSE, for which ChlCast consistently showed lower errors across all forecast lead times.

For independent in situ validation, BGC-Argo surface chlorophyll-a measurements collected during the evaluation period were used as an additional reference. The spatial distribution of the profiles is shown in Fig.~\ref{fig:chlcast_copernicus}c.
When evaluated against BGC-Argo observations, ChlCast had lower MAE and RMSE and a higher Pearson correlation coefficient than Copernicus across all aligned forecast days 1--10 (Fig.~\ref{fig:chlcast_copernicus}d).
When matched profile--forecast pairs were pooled across all ten forecast days, MAE was $0.1385~\mathrm{mg\,m^{-3}}$ for ChlCast and $0.2131~\mathrm{mg\,m^{-3}}$ for Copernicus, a reduction of 35.03\%.
RMSE was $0.3492$ and $0.4017~\mathrm{mg\,m^{-3}}$, respectively, a reduction of 13.07\%.
The Pearson correlation coefficient was 0.6069 for ChlCast, compared with 0.3369 for Copernicus.

To further characterize spatial differences in forecast performance, the evaluation domain was partitioned by ecological province, geographic ocean region, climatological chlorophyll-a concentration and latitude band. Using satellite L4 chlorophyll-a fields as the reference, ChlCast consistently outperformed Copernicus across these groupings, with the largest improvements generally occurring in the mid-latitude Southern Hemisphere. Validation against BGC-Argo observations showed a broadly similar spatial pattern, although the improvements were smaller than those found in the L4-based evaluation (Supplementary Figs.~\ref{fig:supp_l4_regions} and~\ref{fig:supp_argo_regions} and Supplementary Tables~\ref{tab:supp_l4_regions} and~\ref{tab:supp_argo_regions}). Taken together, the satellite L4 and BGC-Argo evaluations showed that ChlCast outperformed the operational Copernicus forecasting system.

\subsection{Representation of fine-scale chlorophyll-a variability}\label{subsec2}

The preceding analyses compared ChlCast with Copernicus at a common spatial resolution of $0.25^\circ$. 
To examine the spatial detail retained at the native 4-km resolution, case-study analyses focused on regions with publicly reported bloom events during the evaluation period. 
For each case, satellite-derived L4 chlorophyll-a fields were compared with ChlCast forecasts at 4-km resolution, the same forecasts aggregated to the $0.25^\circ$ Copernicus grid, and Copernicus forecasts. All forecasts had a lead time of one day.

\begin{figure}[!htbp]
    \centering
    \includegraphics[width=\textwidth]{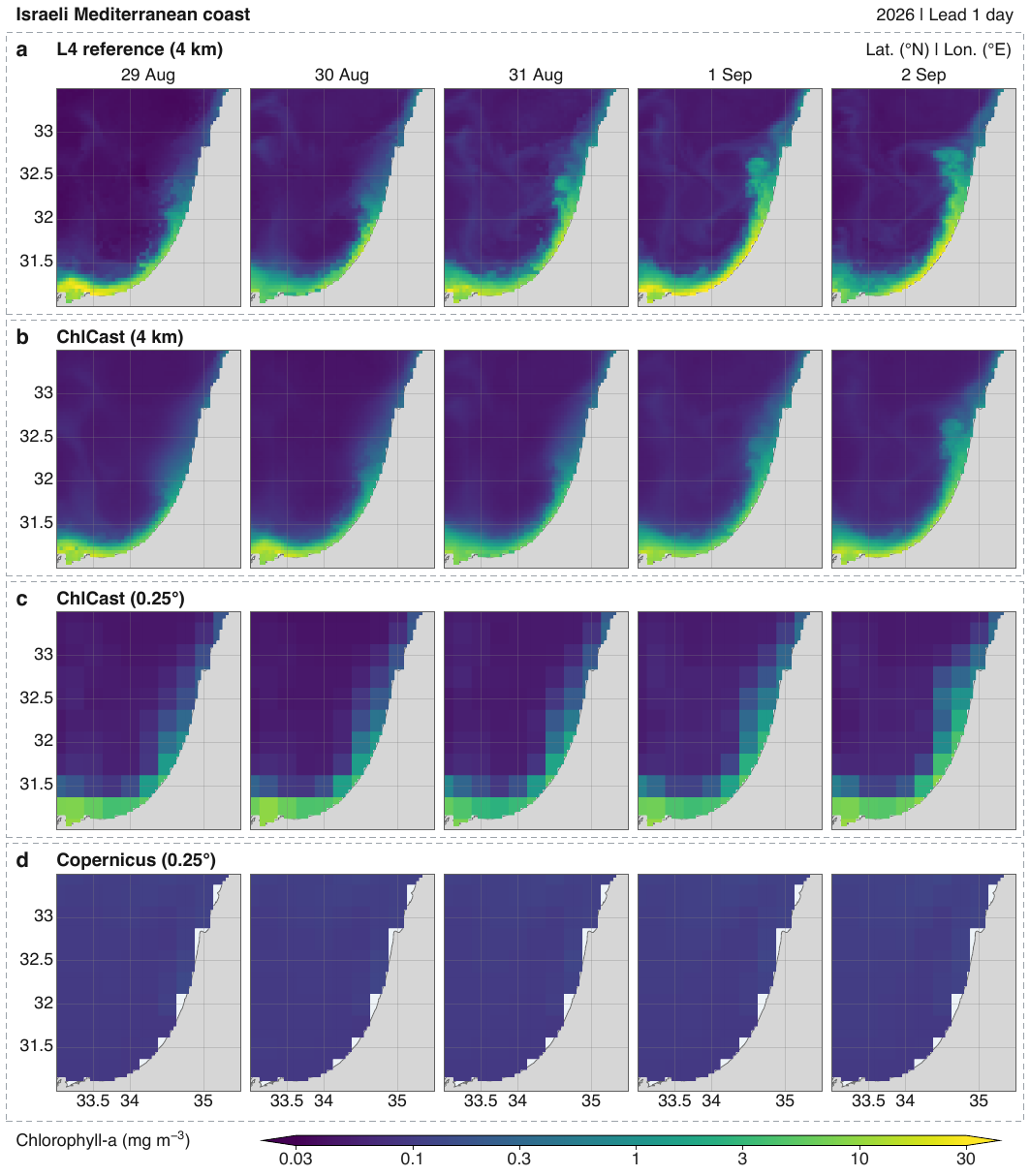}
    \caption{\textbf{Surface chlorophyll-a variability along the Israeli Mediterranean coast from 29 August to 2 September 2026.}
    \textbf{a}, Satellite-derived L4 chlorophyll-a fields at 4-km resolution.
    \textbf{b}, ChlCast 1-day-ahead forecasts at 4-km resolution.
    \textbf{c}, The same ChlCast forecasts aggregated to a $0.25^\circ$ Copernicus grid.
    \textbf{d}, Copernicus 1-day-ahead chlorophyll-a forecasts at $0.25^\circ$ resolution.
    All panels use a common logarithmic colour scale for chlorophyll-a concentration ($\mathrm{mg\,m^{-3}}$) to facilitate comparison of spatial structure and concentration gradients. Dates indicate forecast target dates.
    }
    \label{fig:israel_bloom}
\end{figure}

In late August and early September 2026, an anomalous algal bloom occurred along the Israeli Mediterranean coast, accompanied by increased seawater turbidity. According to a Knesset report issued on 19 September, the event caused five of the country's six major seawater desalination facilities to suspend or reduce production~\cite{knesset2026desalination}. 
From 29 August to 2 September 2026, the satellite L4 fields showed cross-shore chlorophyll-a gradients and alongshore heterogeneity, including narrow high-chlorophyll bands and localized patches whose spatial extent and morphology varied across days (Fig.~\ref{fig:israel_bloom}a). The 4-km ChlCast forecasts captured the coastal high-chlorophyll band and variations among neighbouring coastal sectors (Fig.~\ref{fig:israel_bloom}b). 
When the same ChlCast forecasts were aggregated to the $0.25^\circ$ Copernicus grid, much of this local variability was smoothed and local chlorophyll-a maxima were reduced (Fig.~\ref{fig:israel_bloom}c). Across the 5-day period, the daily spatial maxima in the aggregated forecasts were 32.78--50.47\% lower than those in the 4-km forecasts. Variance decomposition further showed that variability within the $0.25^\circ$ grid cells accounted for 34.26\% of the total spatial variance in the 4-km ChlCast forecasts, compared with 41.27\% in the satellite L4 reference fields. Under the same colour scale, Copernicus forecasts showed weaker spatial contrasts and relatively little change in the overall pattern across the five target dates (Fig.~\ref{fig:israel_bloom}d). Together, these comparisons illustrate the coastal spatial information retained by the native ChlCast forecasts but attenuated by aggregation or less evident in the Copernicus forecasts.

\begin{figure}[!htbp]
    \centering
    \includegraphics[width=\textwidth]{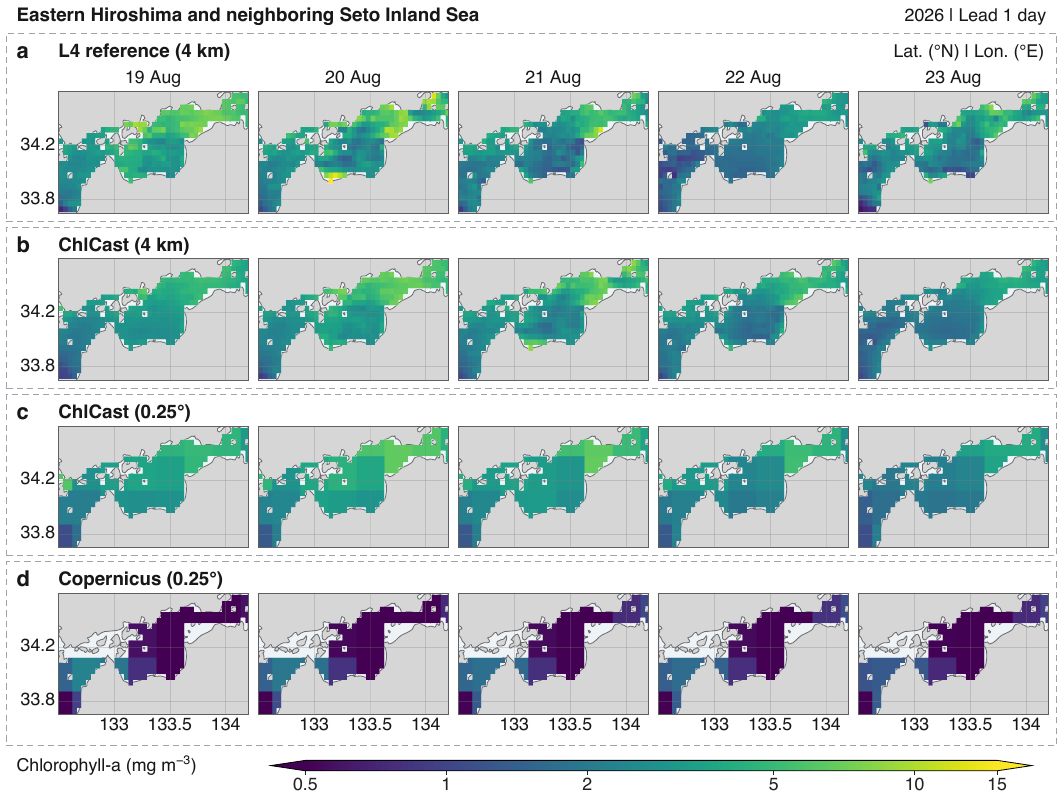}
    \caption{\textbf{Surface chlorophyll-a variability along the eastern coast of Hiroshima Prefecture and in the neighbouring Seto Inland Sea from 19 to 23 August 2026.}
    \textbf{a}, Satellite-derived L4 chlorophyll-a fields at 4-km resolution.
    \textbf{b}, ChlCast 1-day-ahead forecasts at 4-km resolution.
    \textbf{c}, The same ChlCast forecasts aggregated to a $0.25^\circ$ Copernicus grid.
    \textbf{d}, Copernicus 1-day-ahead chlorophyll-a forecasts at $0.25^\circ$ resolution.
    All panels use a common logarithmic colour scale for chlorophyll-a concentration ($\mathrm{mg\,m^{-3}}$) to facilitate comparison of spatial variability among neighbouring coastal waters. Dates indicate forecast target dates.
    }
    \label{fig:hiroshima_bloom}
\end{figure}

A second case was examined along the eastern coast of Hiroshima Prefecture, Japan, where a red-tide advisory was issued on 12 August 2026~\cite{hiroshima2026redtide}. Satellite-derived L4 chlorophyll-a fields from 19 to 23 August were used as the reference (Fig.~\ref{fig:hiroshima_bloom}). The satellite L4 fields showed higher chlorophyll-a concentrations in northern waters and lower concentrations farther south (Fig.~\ref{fig:hiroshima_bloom}a), and the 4-km ChlCast forecasts captured these spatial concentration differences (Fig.~\ref{fig:hiroshima_bloom}b).
After aggregation to the $0.25^\circ$ Copernicus grid, local contrasts in the ChlCast forecasts were reduced (Fig.~\ref{fig:hiroshima_bloom}c), with the maximum chlorophyll-a concentration decreasing by 22.82\% relative to the 4-km forecasts and variability within the $0.25^\circ$ grid cells accounting for 22.89\% of the total spatial variance.
By contrast, the Copernicus forecasts showed little indication of the elevated chlorophyll-a concentrations observed in the satellite L4 fields (Fig.~\ref{fig:hiroshima_bloom}d).

An additional case in the southwestern Baltic Sea during 24--27 August 2026 similarly showed that the 4-km ChlCast forecasts provided finer spatial detail in chlorophyll-a concentration than the corresponding $0.25^\circ$ fields (Supplementary Fig.~\ref{fig:supp_baltic}). These event-based case studies illustrate the value of ChlCast in providing advance information on chlorophyll-a distributions. The forecasts captured major spatial features in the L4 reference fields while retaining concentration gradients and localized maxima that were weakened by coarse-grid averaging. Such spatial detail can support targeted monitoring of chlorophyll-a variability.

\subsection{Evaluation of input configurations and model architectures}\label{subsec3}

To evaluate the design choices underlying ChlCast, the effects of model architecture and data representation were assessed by comparing the convolutional (2+1)D U-Net used in ChlCast with a self-attention-based 3D Swin-T U-Net across three input configurations on the 2023--2025 test set. 
These configurations examined whether representing chlorophyll-a relative to its seasonal background improves forecasting and whether supplying that background provides additional predictive value. M0 used log-transformed chlorophyll-a concentrations as input and predicted future log-transformed concentrations. M1 used log-space anomalies relative to the daily climatological mean as input to predict future anomalies. M2 likewise predicted anomalies but included the corresponding log-transformed daily climatological mean as an additional input channel.

Across both architectures, M2 achieved the lowest MAE and RMSE and the highest anomaly correlation coefficient (ACC) when evaluated across all 10 forecast lead days (Table~\ref{tab:input_architecture_comparison}), indicating improved forecast performance when anomalies and daily climatology were supplied jointly. The Pearson correlation coefficient and log-space errors also generally favoured M2 over M0 and M1 (Supplementary Table~\ref{tab:supp_input_metrics}). 
Under the M2 input configuration, the (2+1)D U-Net and 3D Swin-T U-Net contain 1.79 million and 41.55 million parameters, respectively. Despite having approximately 23.2 times as many parameters as the (2+1)D U-Net, the 3D Swin-T U-Net reduced RMSE by only 0.08\%, while increasing MAE by 0.89\% and decreasing ACC by 0.0007. 
The (2+1)D U-Net with M2 therefore offered a favourable balance between forecast performance and model size and was selected as the final ChlCast configuration. 
On a single NVIDIA GeForce RTX 3090 GPU, ChlCast generated a full-domain 10-day forecast in approximately 34 s. 
This runtime included spatial tiling, data transfer, model inference, stitching and the reconstruction of chlorophyll-a concentrations.

\begin{table}[t]
\centering
\caption{\textbf{Forecast performance across model architectures and input configurations (2023--2025).} MAE and RMSE are expressed in $\mathrm{mg\,m^{-3}}$. Arrows indicate whether lower ($\downarrow$) or higher ($\uparrow$) values represent better performance, and bold values denote the best result for each metric within each architecture.}
\label{tab:input_architecture_comparison}
\begin{tabular}{llrrr}
\toprule
Architecture & Input configuration & MAE $\downarrow$ & RMSE $\downarrow$ & ACC $\uparrow$ \\
\midrule
(2+1)D U-Net & M0 & 0.0687 & 0.5127 & 0.4467 \\
(2+1)D U-Net & M1 & 0.0677 & 0.5014 & 0.4675 \\
(2+1)D U-Net & M2 & \textbf{0.0675} & \textbf{0.5004} & \textbf{0.4709} \\
\midrule
3D Swin-T U-Net & M0 & 0.0697 & 0.5147 & 0.4293 \\
3D Swin-T U-Net & M1 & 0.0782 & 0.5021 & 0.4684 \\
3D Swin-T U-Net & M2 & \textbf{0.0681} & \textbf{0.5000} & \textbf{0.4702} \\
\bottomrule
\end{tabular}
\end{table}

\section{Discussion}
\label{sec:discussion}

This study demonstrates that observation-driven deep learning can provide daily forecasts of surface chlorophyll-a concentration across much of the global ocean at approximately 4-km resolution.
Within the evaluated spatial and temporal scope, ChlCast outperformed the three simple baselines in the satellite L4-based evaluation and achieved lower forecast errors than the operational Copernicus forecasting system when evaluated against both satellite-derived L4 fields and independent BGC-Argo observations.
At 4-km resolution, the forecasts also preserved coastal gradients, localized concentration maxima and alongshore heterogeneity that became less distinct after aggregation to $0.25^\circ$.
ChlCast completed a full-domain 10-day forecast in approximately 34~s on a single NVIDIA GeForce RTX 3090 GPU.
Together, these capabilities suggest that ChlCast could support adaptive oceanographic sampling~\cite{ford2022solution,partridge2026digital} and targeted coastal water-quality monitoring~\cite{zahir2024review,alotaibi2022review}.

Two factors may help explain the forecast skill of ChlCast. First, the preceding 10-day chlorophyll-a sequence contains information about recent spatiotemporal changes, reflecting the combined effects of physical transport and net biological processes that can inform predictions over the subsequent 1--10 days~\cite{martin2003,behrenfeld2018}. 
Second, representing chlorophyll-a as anomalies relative to a daily climatology allows the model to focus on short-term deviations from regionally and seasonally varying background conditions. 
Previous studies have also shown the contributions of seasonal cycles and anomalies to chlorophyll-a variability~\cite{lim2022}. 
Consistent with this interpretation, combining chlorophyll-a anomalies with climatological background information yielded better forecast performance than the other input configurations in the domain-wide evaluation.

The evaluations also revealed an imbalance in forecast errors across climatological chlorophyll-a concentration classes, together with an overall tendency to underestimate chlorophyll-a concentrations.
In the pooled L4-based evaluation, classes with climatological concentrations of at least $1~\mathrm{mg\,m^{-3}}$ accounted for only 2.15\% of the total valid sample weight but contributed 56.79\% of ChlCast's total weighted sum of squared errors (Supplementary Table~\ref{tab:supp_l4_regions}). 
Several factors may contribute to these patterns. 
First, chlorophyll-a concentrations span several orders of magnitude, so comparable relative errors produce larger squared errors at high concentrations~\cite{campbell1995}. 
The concentration of squared errors in productive waters therefore does not necessarily indicate poorer relative predictive performance. 
Second, ChlCast is optimized using a weighted Huber loss in log-anomaly space, whereas forecast errors are evaluated in the original concentration space after exponential back-transformation. Predictions optimized in transformed space do not generally recover the conditional arithmetic mean after back-transformation~\cite{duan1983,sprugel1983}. 
This difference may contribute to the observed negative bias. 
Third, coastal and upwelling waters can experience rapid chlorophyll-a changes associated with episodic forcing, including wind-induced mixing, upwelling pulses and riverine discharge~\cite{cloern1996,chavez2009}. Such externally driven changes may be difficult to anticipate from historical chlorophyll-a sequences alone.

In addition, the comparison between ChlCast and Copernicus should be interpreted in light of differences in native spatial resolution, modelling approaches, information sources and initialization procedures, making it difficult to attribute their performance differences to a single factor.
The satellite L4 product serves as both the training target and the primary gridded evaluation reference, and contains interpolation and retrieval uncertainties in optically complex coastal waters~\cite{garnesson2019,odermatt2012}. BGC-Argo observations provide a complementary reference, but sparse sampling in high-concentration waters limits independent verification under these conditions.
The direct comparison with Copernicus covered only 56 forecast initializations. Relative forecast performance across a complete seasonal cycle therefore remains to be established.
As an observation-driven model, ChlCast depends on the quality and latency of the underlying satellite products for operational use.

Further work on loss-function design to improve forecasts at high chlorophyll-a concentrations and probabilistic methods to quantify forecast uncertainty. Future research could also explore phytoplankton bloom forecasting to support early warning applications and extend AI-based forecasting approaches to three-dimensional ocean biogeochemical fields.

\section{Methods}

\subsection{Datasets}

The Copernicus GlobColour Level-4 global ocean-colour chlorophyll product distributed by the Copernicus Marine Service was used in this study. Daily chlorophyll-a concentration fields from 2008 to 2025 (variable \texttt{CHL}; in $\mathrm{mg\,m^{-3}}$) were used for model training, validation and testing. The study domain covered $60^\circ\mathrm{S}$--$60^\circ\mathrm{N}$ and $180^\circ\mathrm{W}$--$180^\circ\mathrm{E}$ on a regular latitude--longitude grid. The product has a nominal spatial resolution of 4~km, and the selected domain comprised $2880\times8640$ grid cells. The product combines observations from multiple satellite ocean-colour sensors and fills observational gaps through spatiotemporal interpolation~\cite{copernicus_oc_pum}. Daily fields from the same product were also used as the satellite reference for the 2026 forecast evaluation. 

Operational numerical forecasts were obtained from the Copernicus Global Ocean Biogeochemistry Analysis and Forecast product, produced by Mercator Ocean International. Official quality assessments based on satellite and BGC-Argo observations indicate that the system reproduces the large-scale spatial distribution and temporal variability of chlorophyll-a with reasonable fidelity~\cite{copernicus_quality}. The product has a horizontal resolution of $0.25^\circ\times0.25^\circ$ and provides 10-day forecasts updated daily~\cite{copernicus_bgc_pum}. Surface chlorophyll-a concentration was extracted, and 56 valid forecast initializations between 14 August and 9 October 2026 were retained. Forecasts at lead times of 1--10 days were used for comparison with ChlCast.

Independent in situ validation was performed using contemporaneous chlorophyll-a observations from BGC-Argo floats. Adjusted chlorophyll-a concentrations (\texttt{CHLA\_ADJUSTED}) were used, and only measurements with \texttt{CHLA\_ADJUSTED\_QC = 1} were retained~\cite{bittig2019}. For each profile, the shallowest valid observation within 0--10~dbar was selected as the near-surface chlorophyll-a reference. After quality control and spatiotemporal matching with valid forecasts from both systems, the validation dataset comprised 1,915 unique profiles collected from 505 floats between 15 August and 19 October 2026 within $60^\circ\mathrm{S}$--$60^\circ\mathrm{N}$.

\subsection{Data partitioning and preprocessing}

The model-development dataset was partitioned chronologically into training, validation and test periods covering 2008--2019, 2020--2022 and 2023--2025, respectively. Each sample consisted of 10 consecutive days of historical data as input and the subsequent 5 consecutive days as forecast targets. Chlorophyll-a fields from the same product covering 1998--2007 were also used to construct the climatologies.

Daily chlorophyll-a climatologies were constructed separately for each year. For a year $y$, only historical observations from 1998 through $y-1$ were used. At each grid location, a cyclic 31-day calendar window centred on each calendar day, spanning 15 days before and after that day, was used. The climatological concentration was calculated as the arithmetic mean of all valid observations falling within this window across the available historical years. For example, the climatology for 2008 was constructed using observations from 1998--2007, whereas that for 2025 was constructed using observations from 1998--2024.

Chlorophyll-a concentration and the corresponding daily climatology were transformed using the natural logarithm, and the log-space chlorophyll-a anomaly was defined as
\begin{equation}
\begin{aligned}
L(t,\mathbf{x}) &= \ln\!\left[C(t,\mathbf{x})+\varepsilon\right], \\
L_{\mathrm{clim}}(t,\mathbf{x}) &= \ln\!\left[C_{\mathrm{clim}}(t,\mathbf{x})+\varepsilon\right], \\
A(t,\mathbf{x}) &= L(t,\mathbf{x})-L_{\mathrm{clim}}(t,\mathbf{x}),
\end{aligned}
\end{equation}
where $C(t,\mathbf{x})$ and $C_{\mathrm{clim}}(t,\mathbf{x})$ denote the chlorophyll-a concentration and corresponding climatological concentration, respectively, at date $t$ and location $\mathbf{x}$. Concentrations are expressed in $\mathrm{mg\,m^{-3}}$, and $\varepsilon=0.01~\mathrm{mg\,m^{-3}}$ was used to retain valid zero values and avoid taking the logarithm of zero.

For M1 and M2, chlorophyll-a concentrations were reconstructed by adding the predicted log-space anomalies to the log-transformed climatology for each target date:
\begin{equation}
\widehat{C}(t+k,\mathbf{x}) =
\max\!\left\{
\exp\!\left[
\widehat{A}(t+k,\mathbf{x})+
L_{\mathrm{clim}}(t+k,\mathbf{x})
\right]-\varepsilon,\,0
\right\},
\qquad k=1,\ldots,5,
\end{equation}
where $\widehat{A}$ denotes the predicted log-space anomaly, $t$ denotes the final day of the 10-day input sequence and $k$ indexes the 5 target days within each prediction pass. For M0, predicted log-transformed concentrations were converted directly to concentration using the corresponding inverse transformation.

\subsection{Forecast evaluation and baseline methods}

Model architectures and input configurations were evaluated on the 2023--2025 test set using forecasts at lead times of 1--10 days on the 4-km grid. 

For the operational comparison with Copernicus, 56 valid forecast initializations between 14 August and 9 October 2026 were used. ChlCast forecasts and satellite L4 reference fields were aggregated to the $0.25^\circ$ Copernicus grid using area-weighted averaging, and the evaluation was restricted to ocean grid cells between $60^\circ\mathrm{S}$ and $60^\circ\mathrm{N}$.

For BGC-Argo validation, forecast target dates were matched to observation dates in Coordinated Universal Time (UTC). Forecast values were extracted from the nearest grid cell on the common $0.25^\circ$ grid at each float location. Only matched records for which both ChlCast and Copernicus provided valid forecasts were retained.

Forecast performance was evaluated using mean absolute error (MAE), root mean square error (RMSE), mean bias, Pearson correlation coefficient ($r$) and anomaly correlation coefficient (ACC)~\cite{jolliffe2012}. Let $F_n$ and $O_n$ denote the forecast and reference chlorophyll-a concentrations, respectively, $w_n$ the sample weight, and $W=\sum_n w_n$. The error metrics were defined as
\begin{equation}
\begin{aligned}
\mathrm{MAE}
&=
\frac{1}{W}
\sum_n
w_n|F_n-O_n|,\\
\mathrm{RMSE}
&=
\left[
\frac{1}{W}
\sum_n
w_n(F_n-O_n)^2
\right]^{1/2},\\
\mathrm{Bias}
&=
\frac{1}{W}
\sum_n
w_n(F_n-O_n).
\end{aligned}
\end{equation}

For gridded evaluation, the sample weight was defined as
\begin{equation}
w_n=\cos\phi_n,
\end{equation}
where $\phi_n$ is the latitude of grid cell $n$. BGC-Argo observation--forecast pairs were assigned equal weight. The weighted Pearson correlation coefficient was calculated as
\begin{equation}
r
=
\frac{
\sum_n
w_n
(F_n-\overline{F})
(O_n-\overline{O})
}{
\left[
\sum_n
w_n(F_n-\overline{F})^2
\sum_n
w_n(O_n-\overline{O})^2
\right]^{1/2}
},
\end{equation}
where $\overline{F}$ and $\overline{O}$ denote the corresponding weighted means, defined as
\begin{equation}
\overline{F}=\frac{1}{W}\sum_n w_nF_n,
\qquad
\overline{O}=\frac{1}{W}\sum_n w_nO_n.
\end{equation}

ACC was calculated from forecast and reference anomalies relative to the climatology of the target date:
\begin{equation}
a_n^{F}=F_n-C_{\mathrm{clim},n},
\qquad
a_n^{O}=O_n-C_{\mathrm{clim},n}.
\end{equation}
ACC was computed as the weighted centred correlation between $a_n^{F}$ and $a_n^{O}$ using the definition above. The anomalies were defined in the original concentration space. Supplementary log-space MAE and RMSE were calculated from $\ln(F_n+\varepsilon)$ and $\ln(O_n+\varepsilon)$ using the same $\varepsilon$ as in data preprocessing.

All metrics were calculated separately for lead times of 1--10 days. In addition, samples from all ten lead times were pooled to calculate overall ten-day metrics. Pooled metrics were computed directly from the combined weighted samples.

The 2026 operational comparison included three simple baselines: absolute persistence, fixed climatology and linear trend extrapolation. 
Absolute persistence retained the chlorophyll-a concentration on the initialization date for all subsequent lead times. 
The fixed-climatology baseline retained the climatological concentration corresponding to the initialization date for all forecast lead times. 
A linear trend was fitted at each grid cell by ordinary least squares to daily chlorophyll-a concentrations in the 10-day input window ($t=-9,\ldots,0$) and extrapolated to forecast lead times of 1--10 days. Negative extrapolated concentrations were set to zero.

\subsection{Regional assessment}

Regional performance was evaluated using four partitioning schemes based on ecological provinces, geographic ocean regions, climatological chlorophyll-a concentration and latitude. 
Ecological provinces were defined using Longhurst Provinces v4~\cite{longhurst_provinces}, whereas geographic ocean regions were defined using the Natural Earth \texttt{ocean\_basins\_50} dataset~\cite{naturalearth_marine}. 
Both categorical datasets were first mapped to the 4-km ($1/24^\circ$) grid and subsequently aggregated to the common $0.25^\circ$ evaluation grid. 
Each target grid cell was assigned to the category occupying the largest fraction of its ocean area. 
BGC-Argo observation--forecast pairs were assigned according to the category of their matched forecast grid cell. 
Grid cells not assigned to an ecological province or geographic ocean region were excluded from the corresponding regional analyses but retained in the domain-wide statistics.

Climatological chlorophyll-a classes were derived from the daily climatology constructed from observations spanning 1998--2025. 
The arithmetic mean of the 365 daily climatological concentration fields was calculated to obtain a fixed annual-mean climatology. 
This field was then aggregated to the $0.25^\circ$ grid using area-weighted averaging and divided into five classes: $<0.1$, $0.1$--$<0.3$, $0.3$--$<1$, $1$--$<3$ and $\geq3~\mathrm{mg\,m^{-3}}$. 
Latitude was partitioned into four bands: $[-60^\circ,-30^\circ)$, $[-30^\circ,0^\circ)$, $[0^\circ,30^\circ)$ and $[30^\circ,60^\circ]$. 

To quantify the contribution of individual regions to total forecast error and to the error reduction between ChlCast and Copernicus, the weighted sums of absolute and squared errors for model $m$ in region $\mathcal{R}$ were defined as
\begin{equation}
\begin{aligned}
A_{m,\mathcal{R}}
&=
\sum_{n\in\mathcal{R}}
w_n
|F_{m,n}-O_n|,\\
S_{m,\mathcal{R}}
&=
\sum_{n\in\mathcal{R}}
w_n
(F_{m,n}-O_n)^2.
\end{aligned}
\end{equation}

The regional sample-weight fraction $W_{\mathcal{R}}$, absolute-error contribution $P_{m,\mathcal{R}}$, squared-error contribution $Q_{m,\mathcal{R}}$ and contribution to the domain-wide reduction in weighted absolute error $G_{\mathcal{R}}$ were defined as
\begin{equation}
\begin{aligned}
W_{\mathcal{R}}
&=
\frac{
\sum_{n\in\mathcal{R}}w_n
}{
\sum_n w_n
}
\times100\%,\\[4pt]
P_{m,\mathcal{R}}
&=
\frac{
A_{m,\mathcal{R}}
}{
A_{m,\mathrm{all}}
}
\times100\%,\\[4pt]
Q_{m,\mathcal{R}}
&=
\frac{
S_{m,\mathcal{R}}
}{
S_{m,\mathrm{all}}
}
\times100\%,\\[4pt]
G_{\mathcal{R}}
&=
\frac{
A_{\mathrm{Copernicus},\mathcal{R}}
-
A_{\mathrm{ChlCast},\mathcal{R}}
}{
A_{\mathrm{Copernicus},\mathrm{all}}
-
A_{\mathrm{ChlCast},\mathrm{all}}
}
\times100\%.
\end{aligned}
\end{equation}
Here, $\mathrm{all}$ denotes all samples across the evaluation domain. All regional contribution statistics were calculated from samples pooled across the ten forecast lead times.

\subsection{Analysis of local spatial structure}

Three case-study regions were selected to assess local spatial structure. The Israeli Mediterranean coast case covered $31^\circ$--$33.5^\circ\mathrm{N}$ and $33^\circ$--$35.5^\circ\mathrm{E}$ from 29 August to 2 September 2026. The eastern Hiroshima Prefecture and adjacent Seto Inland Sea case covered $33.7^\circ$--$34.6^\circ\mathrm{N}$ and $132.5^\circ$--$134.2^\circ\mathrm{E}$ from 19 to 23 August 2026. The southwestern Baltic Sea case covered $54^\circ$--$56^\circ\mathrm{N}$ and $12^\circ$--$16^\circ\mathrm{E}$ from 24 to 27 August 2026. For each target date, the 1-day-ahead forecast initialized on the preceding day was analysed. Satellite L4 reference fields, 4-km ChlCast forecasts, the same ChlCast forecasts aggregated to $0.25^\circ$, and corresponding 1-day-ahead Copernicus forecasts were compared. Coarse-grid chlorophyll-a concentration was calculated as the area-weighted mean of source-grid ocean values, and the 4-km ChlCast forecasts and their $0.25^\circ$ aggregates were compared over the same ocean domain. Within each case study, all forecast and reference fields were visualized using a common logarithmic colour scale, whereas quantitative analyses were performed using chlorophyll-a concentration in the original linear space.

To quantify spatial variability occurring within coarse grid cells, 4-km ocean grid cells within each study region were grouped according to their corresponding $0.25^\circ$ grid cell. Let $C_{d,i}$ denote chlorophyll-a concentration on date $d$ at fine-grid cell $i$, $w_i=\cos\phi_i$ the corresponding area weight, $\mu_{d,g}$ the weighted mean concentration within coarse-grid cell $g$ on date $d$, and $\mu_d$ the weighted mean concentration over the full study region on that date. The fraction of total spatial variance attributable to within-cell variability was defined as
\begin{equation}
R_{\mathrm{within}}
=
\frac{
\displaystyle
\sum_d
\sum_g
\sum_{i\in\Omega_{d,g}}
w_i
\left(
C_{d,i}-\mu_{d,g}
\right)^2
}{
\displaystyle
\sum_d
\sum_{i\in\Omega_d}
w_i
\left(
C_{d,i}-\mu_d
\right)^2
}
\times100\%,
\end{equation}
where $\Omega_d$ denotes the set of fine-grid ocean cells within the study region on date $d$, and $\Omega_{d,g}$ denotes the subset belonging to coarse-grid cell $g$. The ratio was calculated separately for ChlCast and the satellite L4 reference fields using the same ocean grid cells, with the within-cell and total sums of squares accumulated across all study dates before calculating the ratio.

Daily spatial maximum chlorophyll-a concentrations were also calculated for the 4-km ChlCast forecasts and their $0.25^\circ$ aggregates over the same ocean domain. The relative reduction in the daily spatial maximum caused by spatial aggregation was defined as
\begin{equation}
R_{\mathrm{max},d}
=
\frac{
C_{\mathrm{max},d}^{4\,\mathrm{km}}
-
C_{\mathrm{max},d}^{0.25^\circ}
}{
C_{\mathrm{max},d}^{4\,\mathrm{km}}
}
\times100\%,
\end{equation}
where $C_{\mathrm{max},d}^{4\,\mathrm{km}}$ and $C_{\mathrm{max},d}^{0.25^\circ}$ denote the spatial maxima of the 4-km forecasts and their $0.25^\circ$ aggregates, respectively, on date $d$.

\section*{Data availability}

The satellite-derived chlorophyll-a data used in this study were obtained from the Copernicus Marine Service (product ID: OCEANCOLOUR\_GLO\_BGC\_L4\_MY\_009\_104; \url{https://doi.org/10.48670/moi-00281}).
The daily chlorophyll-a climatology used in this study was calculated from the satellite-derived chlorophyll-a data.
The operational ocean biogeochemical forecasts used for comparison were obtained from the Copernicus Marine Service (product ID: GLOBAL\_ANALYSISFORECAST\_BGC\_001\_028; \url{https://doi.org/10.48670/moi-00015}).
The BGC-Argo chlorophyll-a observations used for independent validation are publicly available through the Argo Global Data Assembly Centres (\url{https://doi.org/10.17882/42182}).

\section*{Code availability}

The code for ChlCast is publicly available on GitHub at \url{https://github.com/hohosilent-hoho/ChlCast}.

\bibliography{sn-bibliography}

\clearpage
\setcounter{figure}{0}
\setcounter{table}{0}
\renewcommand{\figurename}{Supplementary Fig.}
\renewcommand{\tablename}{Supplementary Table}
\renewcommand{\theHfigure}{supp.\arabic{figure}}
\renewcommand{\theHtable}{supp.\arabic{table}}
\section*{Supplementary Information}

\begin{figure}[!ht]
\centering
\includegraphics[width=0.90\textwidth,height=0.75\textheight,keepaspectratio]{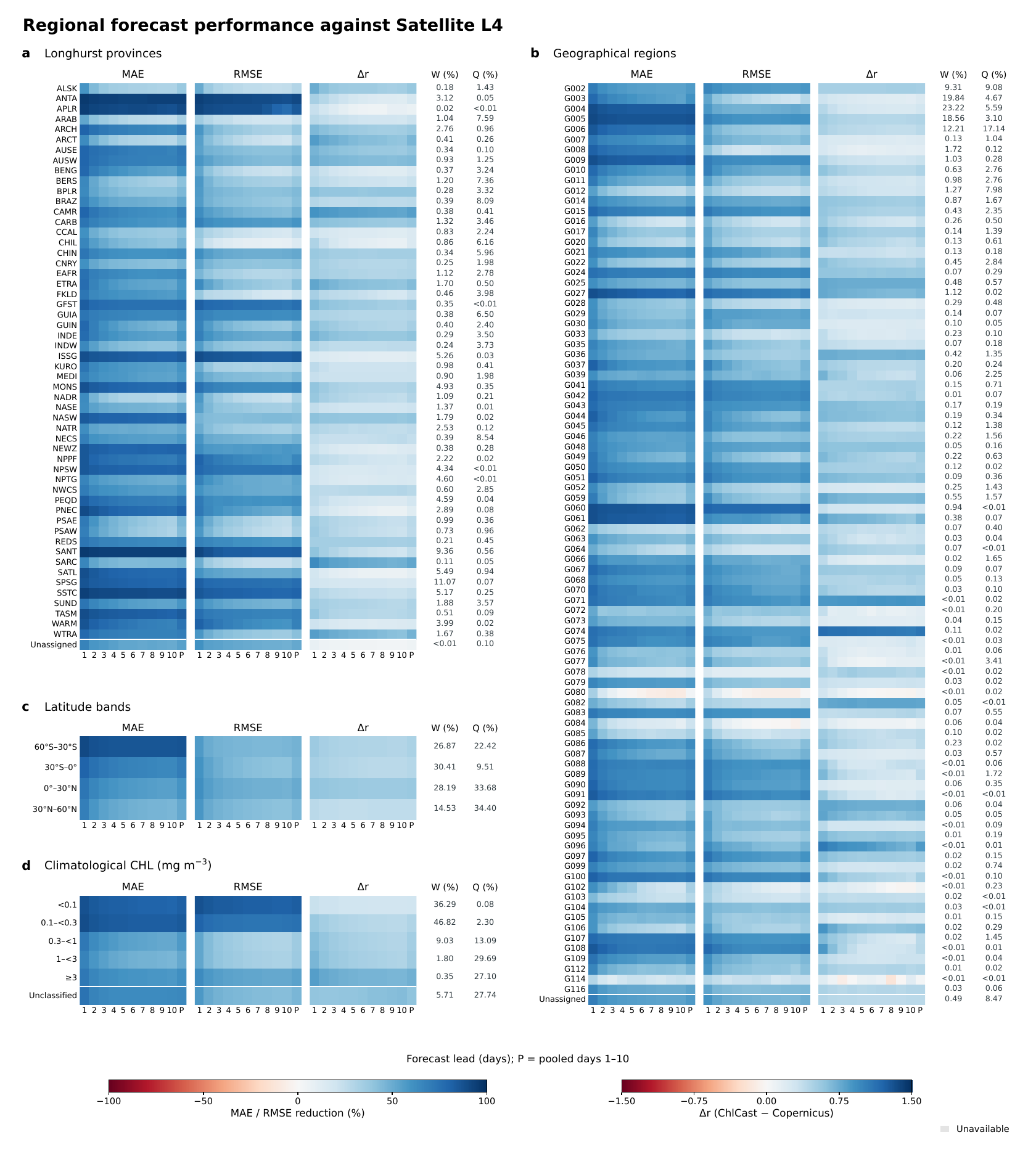}
\caption{\textbf{Regional forecast performance relative to satellite-derived L4 chlorophyll fields.}
ChlCast and Copernicus forecasts are compared over $60^\circ$S--$60^\circ$N on a common $0.25^\circ$ grid, using 56 available initializations from 14 August to 9 October 2026.
\textbf{a}, Longhurst provinces. \textbf{b}, Geographical regions. \textbf{c}, Latitude bands. \textbf{d}, Annual-mean climatological chlorophyll classes ($\mathrm{mg\,m^{-3}}$).
Columns show lead days 1--10 and the pooled result (P). MAE and RMSE denote mean absolute error and root mean square error; their reductions are relative to Copernicus, with positive values indicating lower ChlCast errors. $\Delta r$ is the ChlCast minus Copernicus Pearson correlation coefficient. Blue and red indicate positive and negative values; grey indicates unavailable data or undefined correlation.
W is the percentage of valid sample area weight and Q is the percentage contribution to ChlCast weighted squared error, both pooled over days 1--10. Positive percentages below 0.01 are labelled $<0.01$.
Only sampled provinces and geographical regions are plotted. Unassigned and Unclassified denote unassigned regions and incomplete or missing climatological coverage, respectively. Region names and values are provided in Supplementary Table~\ref{tab:supp_l4_regions}.}
\label{fig:supp_l4_regions}
\end{figure}

\clearpage

\begin{figure}[!ht]
\centering
\includegraphics[width=\textwidth,height=0.80\textheight,keepaspectratio]{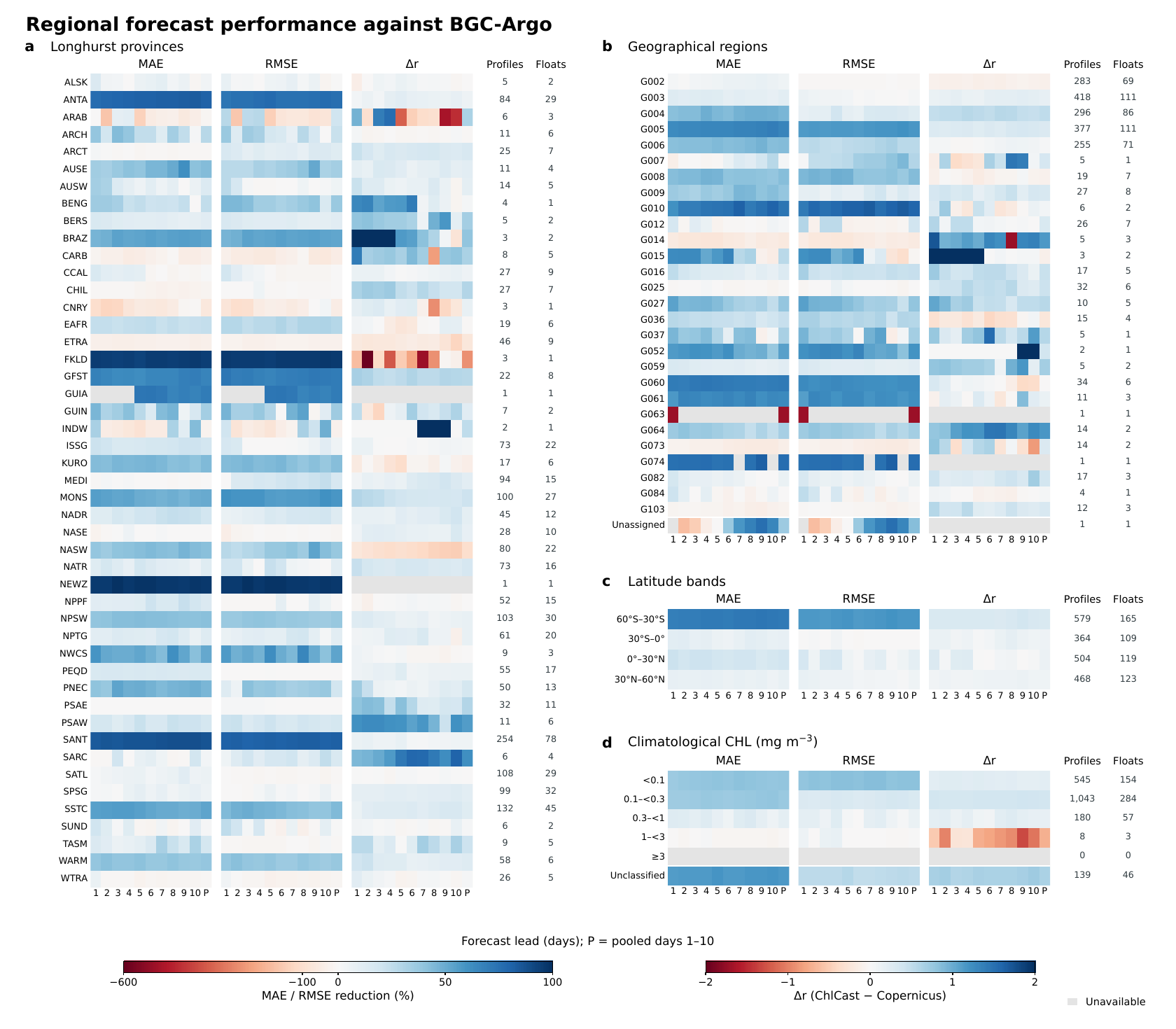}
\caption{\textbf{Regional forecast performance relative to BGC-Argo chlorophyll observations.}
ChlCast and Copernicus forecasts are evaluated over $60^\circ$S--$60^\circ$N against biogeochemical Argo (BGC-Argo) observations from 15 August to 19 October 2026, using the same 16,180 equally weighted profile--forecast pairs.
\textbf{a}, Longhurst provinces. \textbf{b}, Geographical regions. \textbf{c}, Latitude bands. \textbf{d}, Annual-mean climatological chlorophyll classes ($\mathrm{mg\,m^{-3}}$).
Columns show lead days 1--10 and the pooled result (P). MAE and RMSE denote mean absolute error and root mean square error; their reductions are relative to Copernicus, with positive values indicating lower ChlCast errors. $\Delta r$ is the ChlCast minus Copernicus Pearson correlation coefficient. Blue and red indicate positive and negative values; grey indicates unavailable data or undefined correlation.
Profiles and Floats give the numbers of distinct profiles and floats in the pooled samples; a profile can verify multiple lead times. Only sampled provinces and geographical regions are plotted. Unassigned and Unclassified denote unassigned regions and incomplete or missing climatological coverage, respectively. Region names, unsampled categories and full results are provided in Supplementary Table~\ref{tab:supp_argo_regions}.}
\label{fig:supp_argo_regions}
\end{figure}

\clearpage

\begin{figure}[!ht]
\centering
\includegraphics[width=\textwidth,height=0.80\textheight,keepaspectratio]{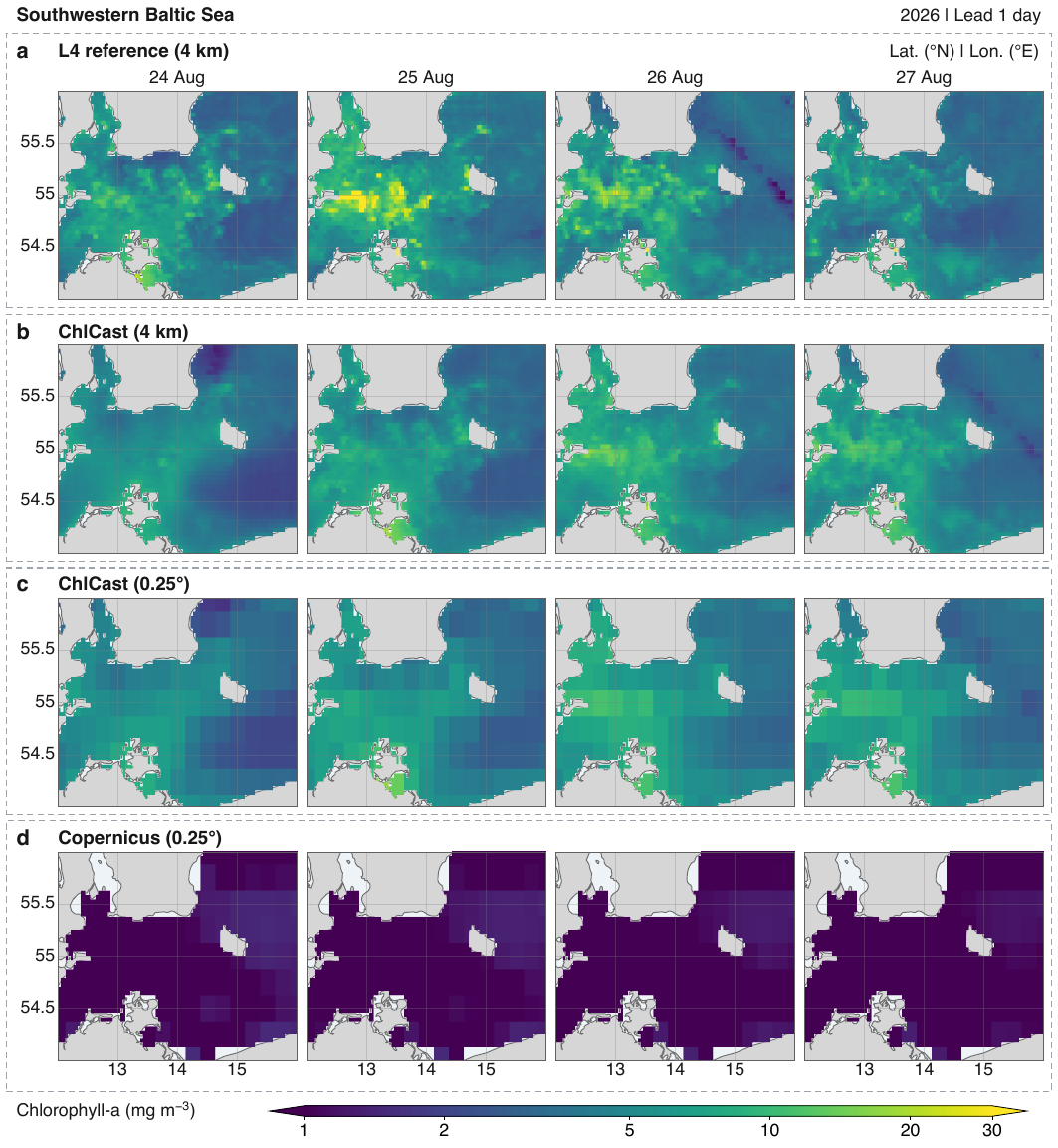}
\caption{\textbf{Chlorophyll fields and 1-day-ahead forecasts in the southwestern Baltic Sea.}
The domain is $54^\circ$--$56^\circ$N, $12^\circ$--$16^\circ$E. Columns correspond to 24, 25, 26 and 27 August 2026.
\textbf{a}, Satellite-derived L4 chlorophyll reference fields at 4-km resolution. \textbf{b}, 4-km ChlCast forecasts. \textbf{c}, The same ChlCast forecasts aggregated to a $0.25^\circ$ grid. \textbf{d}, Copernicus forecasts at $0.25^\circ$ resolution. Each forecast was initialized on the preceding day.
All panels share a logarithmic colour scale of 1--30~$\mathrm{mg\,m^{-3}}$; values outside this range use the endpoint colours. The 4-km ChlCast fields and their $0.25^\circ$ aggregates use the same valid marine display extent. Coarse-grid values are constant within each cell, without spatial interpolation or smoothing. Copernicus retains its own valid coverage; missing values are not filled. Grey denotes land or excluded locations, and pale blue denotes ocean locations without a valid displayed value.}
\label{fig:supp_baltic}
\end{figure}

\begin{table}[!ht]
\centering
\caption{\textbf{Complete regional forecast evaluation against satellite-derived L4 chlorophyll fields.}
ChlCast and Copernicus results over $60^\circ$S--$60^\circ$N on the common $0.25^\circ$ grid, for lead days 1--10 and their pooled result. The 56 available initialization dates span 14 August to 9 October 2026, with target dates from 15 August to 19 October. The electronic table contains four worksheets: Longhurst, Geographical, Latitude and Climatology.
MAE, RMSE and bias are expressed in $\mathrm{mg\,m^{-3}}$; bias is forecast minus reference. MAE and RMSE reductions are relative to Copernicus, with positive values indicating lower ChlCast errors. $r$ and $\Delta r$ are dimensionless, and $\Delta r$ is ChlCast minus Copernicus. Sample weight share is the fraction of valid area weight; Valid grid cells counts distinct evaluated cells. AE and SE contributions are the regional fractions of each model's total absolute and squared error. AE reduction contribution is the regional contribution to the domain-wide reduction in absolute error. Plot ID corresponds to Supplementary Fig.~\ref{fig:supp_l4_regions}. Pooled 1--10 denotes the pooled result and ALL the full evaluation domain. Unassigned, Unclassified and unsampled categories are retained. Unavailable metrics are blank, with No data identifying unsampled categories. Bold rows identify pooled results only.}
\label{tab:supp_l4_regions}
\begin{tabular}{p{0.94\textwidth}}
Complete electronic table: \texttt{supplementary\_table\_2-10.xlsx}. \\
\end{tabular}
\end{table}

\begin{table}[!ht]
\centering
\caption{\textbf{Complete regional forecast evaluation against BGC-Argo chlorophyll observations.}
ChlCast and Copernicus results over $60^\circ$S--$60^\circ$N for lead days 1--10 and their pooled result, using BGC-Argo observations from 15 August to 19 October 2026. The common sample comprises 16,180 equally weighted profile--forecast pairs from 1,915 distinct profiles and 505 floats. The electronic table contains Longhurst, Geographical, Latitude and Climatology worksheets.
MAE, RMSE and bias are expressed in $\mathrm{mg\,m^{-3}}$; bias is forecast minus observation. MAE and RMSE reductions are relative to Copernicus, with positive values indicating lower ChlCast errors. $\Delta r$ is the ChlCast minus Copernicus Pearson correlation coefficient. Matched pairs and Pair share give the pair count and its fraction of the full sample. Distinct profiles, Distinct floats, Target dates and Forecast cases report sample coverage. Profiles can verify multiple lead times, and distinct profile or float counts need not sum across regions. AE and SE contributions are the regional fractions of each model's total absolute and squared error; AE reduction contribution is the regional contribution to the domain-wide reduction in absolute error. Plot ID corresponds to Supplementary Fig.~\ref{fig:supp_argo_regions}. ALL denotes the full domain. Unassigned, Unclassified and unsampled categories are retained. Unavailable metrics are blank; No data identifies unsampled categories. Bold rows identify pooled results only.}
\label{tab:supp_argo_regions}
\begin{tabular}{p{0.94\textwidth}}
Complete electronic table: \texttt{supplementary\_table\_3-10.xlsx}. \\
\end{tabular}
\end{table}

\clearpage
\begin{table*}[!ht]
\centering
\caption{\textbf{Complete forecast evaluation across model architectures and input configurations.}
Results for (2+1)D U-Net and 3D Swin-T U-Net under input configurations M0, M1 and M2, evaluated over $60^\circ\mathrm{S}$--$60^\circ\mathrm{N}$ for target dates from 1 January 2023 to 31 December 2025.
The test set contains 1,087 complete 10-day windows. For each input configuration, all metrics are calculated from data pooled across forecast lead days 1--10.
Bias, mean absolute error (MAE) and root mean square error (RMSE) are expressed in $\mathrm{mg\,m^{-3}}$; bias is forecast minus reference. $r$ is the concentration-space Pearson correlation coefficient, and ACC is the anomaly correlation coefficient. Log MAE and Log RMSE are natural-log-space errors.}
\label{tab:supp_input_metrics}
\fontsize{8}{10}\selectfont
\setlength{\tabcolsep}{4pt}
\renewcommand{\arraystretch}{1.15}
\resizebox{\textwidth}{!}{%
\begin{tabular}{llrrrrrrr}
\toprule
Architecture & Input & Bias & $r$ & Log MAE & Log RMSE & MAE & RMSE & ACC \\
\midrule
(2+1)D U-Net & M0 & -0.0234 & 0.8284 & 0.1558 & 0.2373 & 0.0687 & 0.5127 & 0.4467 \\
(2+1)D U-Net & M1 & -0.0217 & 0.8329 & 0.1540 & 0.2339 & 0.0677 & 0.5014 & 0.4675 \\
(2+1)D U-Net & M2 & -0.0198 & 0.8328 & 0.1529 & 0.2323 & 0.0675 & 0.5004 & 0.4709 \\
\midrule
3D Swin-T U-Net & M0 & -0.0209 & 0.8213 & 0.1571 & 0.2392 & 0.0697 & 0.5147 & 0.4293 \\
3D Swin-T U-Net & M1 & -0.0337 & 0.8325 & 0.2989 & 0.4520 & 0.0782 & 0.5021 & 0.4684 \\
3D Swin-T U-Net & M2 & -0.0198 & 0.8339 & 0.1553 & 0.2345 & 0.0681 & 0.5000 & 0.4702 \\
\bottomrule
\end{tabular}}
\end{table*}
\clearpage

\end{document}